\documentclass[runningheads]{llncs}
\usepackage{graphicx}
\usepackage{amsmath,amssymb} % define this before the line numbering.
\usepackage{color}
\usepackage{multirow}
\usepackage{booktabs}
\usepackage{mathtools}
\usepackage{subfigure}
\newcommand{\etal}{\textit{et al}. }
\newcommand{\ie}{\textit{i}.\textit{e}. }
\newcommand{\eg}{\textit{e}.\textit{g}. }
\newcommand{\nj}[1]{\textcolor{black}{#1}}
\newcommand{\mg}[1]{\textcolor{black}{#1}}
\newcommand{\sj}[1]{\textcolor{black}{#1}}

\newcommand\blfootnote[1]{%
  \begingroup
  \renewcommand\thefootnote{}\footnote{#1}%
  \addtocounter{footnote}{-1}%
  \endgroup
}
\begin{document}
\title{Motion Feature Network: Fixed Motion Filter for Action Recognition} 
% Replace with your title

\titlerunning{Motion Feature Network}
% Replace with a meaningful short version of your title
%
% \author{Seungeui Lee\inst{1} \and
% Myunggi Lee\inst{1,2} \and
% Sungjoon Son\inst{1,2} \and 
% Gyutae Park\inst{1,2} \and \\
% Nojun Kwak\inst{1}}
\author{Myunggi Lee\inst{1,2,*}%\orcidID{1111-2222-3333-4444} 
\and
Seungeui Lee\inst{1,*}%\orcidID{0000-0003-3560-785X} 
\and
Sungjoon Son\inst{1,2}%\orcidID{2222--3333-4444-5555} 
\and 
Gyutae Park\inst{1,2}%\orcidID{2222--3333-4444-5555} 
\and \\ 
Nojun Kwak\inst{1}}%\orcidID{2222--3333-4444-5555}}
%
%Please write out author names in full in the paper, i.e. full given and family names. 
%If any authors have names that can be parsed into FirstName LastName in multiple ways, please include the correct parsing, in a comment to the volume editors:
%\index{Lastnames, Firstnames}
%(Do not uncomment it, because you may introduce extra index items if you do that, we will use scripts for introducing index entries...)
\authorrunning{M. Lee, S. Lee, S. Son, G. Park and N. Kwak}
% Replace with shorter version of the author list. If there are more authors than fits a line, please use A. Author et al.
%

\institute{Seoul National University, Seoul, South Korea \\ \email{\{myunggi89, dehlix, sjson, pgt4861, nojunk\}@snu.ac.kr} \and
V.DO Inc., Suwon, Korea}
\maketitle              % typeset the header of the contribution
\begin{abstract}
Spatio-temporal representations in frame sequences play an important role in the task of action recognition. Previously, a method of using optical flow as a temporal information in combination with a set of RGB images that contain spatial information has shown great performance enhancement in the action recognition tasks. However, it has an expensive computational cost and requires two-stream (RGB and optical flow) framework. In this paper, we propose MFNet (Motion Feature Network) containing motion blocks which make it possible to encode spatio-temporal information between adjacent frames in a unified network that can be trained end-to-end. The motion block can be attached to any existing CNN-based action recognition frameworks with only a small additional cost. We evaluated our network on two of the action recognition datasets (Jester and Something-Something) and achieved competitive performances for both datasets by training the networks from scratch. 
\end{abstract}
%
%
%
% 1. Introduction --------------------------------------------------
\section{Introduction}
\label{sec:intro}
\blfootnote{* M. Lee and S. Lee equally contributed the paper. %as indicated by $\star$.% indicates equal contribution. 
% The order was determined by rolling a dice.
%} \blfootnote{
This work was supported by the ICT R\&D program of MSIP/IITP, Korean Government (2017-0-00306)}Convolutional neural networks (CNNs) \cite{lecun1995convolutional} are originally designed to represent static appearances of visual scenes well.
However, it has a limitation if the underlying structure is characterized by sequential and temporal relations. In particular, since recognizing human behavior in a video requires both spatial appearance and temporal motion as important cues, many previous researches have utilized various modalities that can capture motion information such as optical flow \cite{wang2016temporal} and RGBdiff (temporal difference in consecutive RGB frames) \cite{wang2016temporal}. Methods based on two-stream \cite{feichtenhofer2016convolutional,ng2015beyond,wang2016temporal} and 3D convolutions \cite{carreira2017quo,tran2015learning} utilizing these input modalities achieve state-of-the-art performances in the field of action recognition. 
However, even though optical flow is a widely utilized modality that provides short-term temporal information, it takes a lot of time to generate. Likewise, 3D-kernel-based methods such as 3D ConvNets also require heavy computational burden with high memory requirements. 

% Dataset
\begin{figure}[t]
\centering
	\label{figure:datasets}    \includegraphics[width=\linewidth]{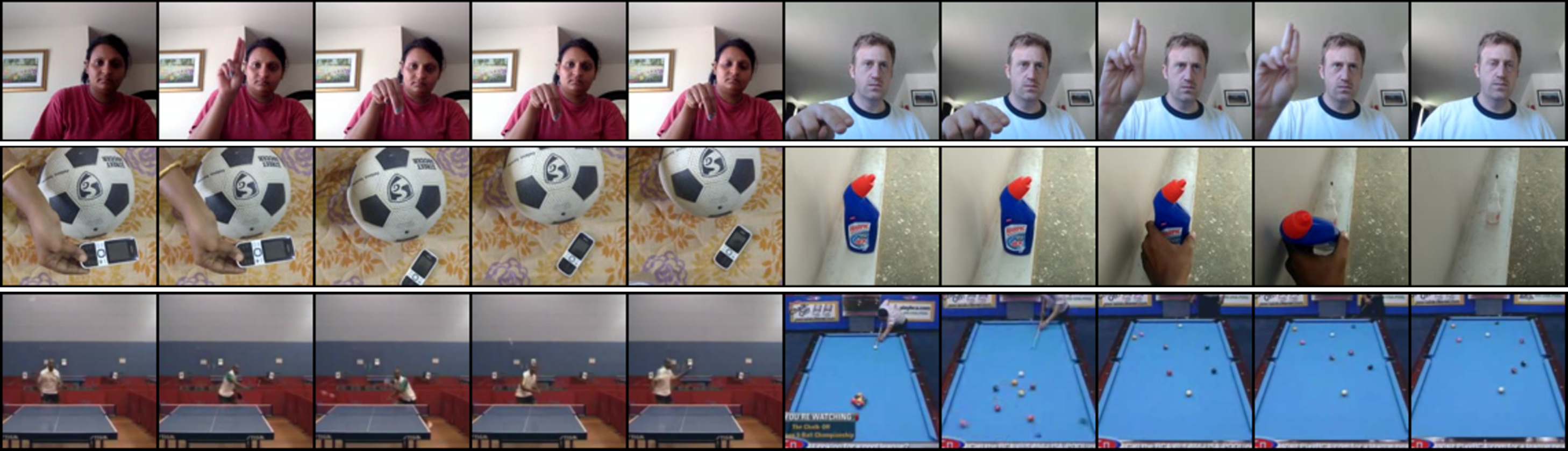}
    \caption{Some examples of action classes in the three action recognition datasets, Jester (top), Something-Something (middle), and UCF101 (bottom). -- top: left \textit{`Sliding Two Fingers Down'}, right \textit{`Sliding Two Fingers Up'}, middle: left \textit{`Dropping something in front of something'}, right \textit{`Removing something, revealing something behind'}, bottom: left \textit{`TableTennisShot'}, right \textit{`Billiards'}. 
    Due to ambiguity of symmetrical pair classes/actions, \nj{static images only} are not enough to recognize correct labels without sequential information \nj{in the former two datasets}. However, in case of the bottom UCF101 image frames, the action class can be recognized with only spatial context (\eg background and objects) from a single image.} 
\end{figure}

In our view, most previous \nj{labeled} action recognition datasets \nj{such as UCF101 \cite{soomro2012ucf101}, HMDB51~\cite{kuehne2013hmdb51}, Sports-1M~\cite{karpathy2014large} and THUMOS~\cite{jiang2014thumos}} provide highly abstract concepts of human behavior. 
Therefore \nj{they} can be \nj{mostly} recognized without \nj{the help of} temporal relations of sequential frames. \nj{For example, the \textit{`Billiard'} and \textit{`TableTennisShot'} in UCF101 can be easily recognizable by just seeing one frame as shown in the third row of Fig. \ref{figure:datasets}.}
Unlike these datasets, Jester \cite{dataset_jester} and Something-Something \cite{goyal2017something} include \nj{more} detailed physical aspects of actions and scenes. The appearance information \nj{has a very limited usefulness in classifying actions for these datasets.}
Also, \nj{visual objects in the scenes} that mainly provide shape information are less important \nj{for the purpose of recognizing actions on these datasets}. In particular, the Something-Something dataset has little correlation between the object and the action class, as its name implies. The first two rows of Fig.~\ref{figure:datasets} show some examples of \nj{these} datasets. As shown in Figure~\ref{figure:datasets}, it is difficult to classify the action class with only one image. Also, even if there are multiple images, the action class can be changed according to the temporal order. Thus, it can be easily confused when using the conventional static feature extractors. Therefore, the ability to extract the temporal relationship between consecutive frames is important to classify human behavior in these datasets. 

To solve these issues, we introduce a unified model which \nj{is named as the} Motion Feature Network (MFNet). MFNet contains specially designed motion blocks which represent spatio-temporal relationships from only RGB frames. Because it extracts temporal information using only RGB, pre-computation time \nj{that is typically needed to compute optical flow} is not needed compared with the existing optical flow-based \nj{approaches}. \nj{Also, because MFNet is} based on a 2D CNN architecture, it has fewer parameters \nj{compared to its 3D counterparts}.

We perform experiments to verify our model's ability to extract spatio-temporal features on \nj{a couple of publicly available action recognition datasets}. In these datasets, each video label is closely related to the sequential relationships among frames. MFNet trained using only RGB frames significantly outperforms previous methods. Thus, MFNet can be used as a good solution for \nj{an} action classification task in videos consisting of sequential relationships of detailed physical entities. We also conduct ablation studies to understand properties of MFNets \nj{in more detail}. 

The rest of this paper is organized as follows. Some related works for action recognition tasks are discussed in Section~\ref{sec:rel}. Then in Section~\ref{sec:model}, we introduce our proposed MFNet architecture in detail. After that, experimental results with ablation studies are presented and analyzed in Section~\ref{sec:exp}. Finally, the paper is concluded in Section~\ref{sec:con}. 

% 2. Related Works --------------------------------------------------
\section{Related Works}
\label{sec:rel}

With the great success of \nj{CNNs} on various computer vision tasks, a growing number of studies have tried to utilize deeply learned features for action recognition in video \nj{datasets}. Especially, as the consecutive frames of input data imply sequential contexts, temporal information as well as spatial information is an important cue for classification tasks.
There have been several approaches to extract these spatio-temporal features on action recognition problems.

One popular way to learn spatio-temporal features is using 3D convolution and 3D pooling hierarchically~\cite{feichtenhofer2016spatiotemporal,hara2017learning,tran2015learning,tran2017convnet,zhang2016real}. \nj{In this approach, they} usually stack continuous frames of \nj{a} video clip and feed them into the network. The 3D convolutions have \nj{enough} capacity to encode spatio-temporal information on densely sampled frames but are inefficient in terms of computational cost. Furthermore, the number of parameters to be optimized are relatively large compared to other approaches. \nj{Thus, it is} difficult to train on small datasets, such as UCF101 \cite{soomro2012ucf101} and HMDB51 \cite{Kuehne11}.
In order to overcome these issues, Carreira \etal \cite{carreira2017quo} introduced a new large dataset named Kinetics\cite{kay2017kinetics}, which facilitates training 3D models. They also suggest inflating 3D convolution filters from 2D convolution filters to bootstrap parameters from the pre-trained ImageNet \cite{deng2009imagenet} models. It achieves state-of-the-art performances in action recognition tasks.

Another famous approach is \nj{the} two-stream-based method proposed by Simonyan \etal \cite{simonyan2014two}. It encodes two kinds of modalities which are raw pixels of \nj{an} image and \nj{the} optical flow extracted from two consecutive raw image frames. It predicts action classes by averaging the predictions from both a single RGB frame and a stack of externally computed multiple optical flow frames. A large amount of follow up studies\cite{miech2017learnable,wang2017appearance,wu2015fusing} to improve the performance of action recognition has been proposed based on the two-stream framework \cite{feichtenhofer2016convolutional,ng2015beyond,wang2016temporal}. 
As an extension to the previous two-stream method, Wang \etal \cite{wang2016temporal} \nj{proposed the} temporal segment network. It samples image frames and optical flow frames on different time segments over the entire video sequences instead of short snippets, \nj{then it} trains RGB frames and optical flow frames independently. \nj{At} inference time, it accumulates the results to predict \nj{an activity class}.
While it brings a significant improvement over traditional methods \cite{dalal2006human,wang2011action,wang2013action}, it still relies on pre-computed optical \nj{flows which are} computationally expensive. 

In order to replace the role of hand-crafted optical flow, there have been some works feeding frames similar to optical flow as inputs to the convolutional networks \cite{wang2016temporal,zhang2016real}. Another line of works use optical flow only in training phase as ground-truth \cite{ng2016actionflownet,zhu2017hidden}. They trained \nj{a} network \nj{that reconstructs} optical flow images \nj{from raw images and provide the estimated optical flow information to the action recognition network.} 
Recently, Sun \etal \cite{DBLP:journals/corr/abs-1711-11152} proposed \nj{a method of optical-flow-guided features}. It extracts motion representation using two \nj{sets of features from adjacent frames} by \nj{separately applying temporal subtraction (temporal features)  and Sobel filters (spatial features).} Our proposed method is highly related to this work. The differences are that we \sj{feedforward} spatial and temporal features in a unified network instead of separating two features apart. Thus, it is \nj{possible to train the proposed MFNet} in an end-to-end manner.

% 3. Model --------------------------------------------------
\section{Model}
\label{sec:model}

In this section, we first introduce the overall architecture of \nj{the proposed MFNet} and then give a detailed description \nj{of `motion filter' and `motion block'} \nj{which constitute MFNet}. We provide several instantiations of motion filter and motion block to explain the intuition behind it. 

\begin{figure}[t]
	\centering
    \includegraphics[width=.85\linewidth]{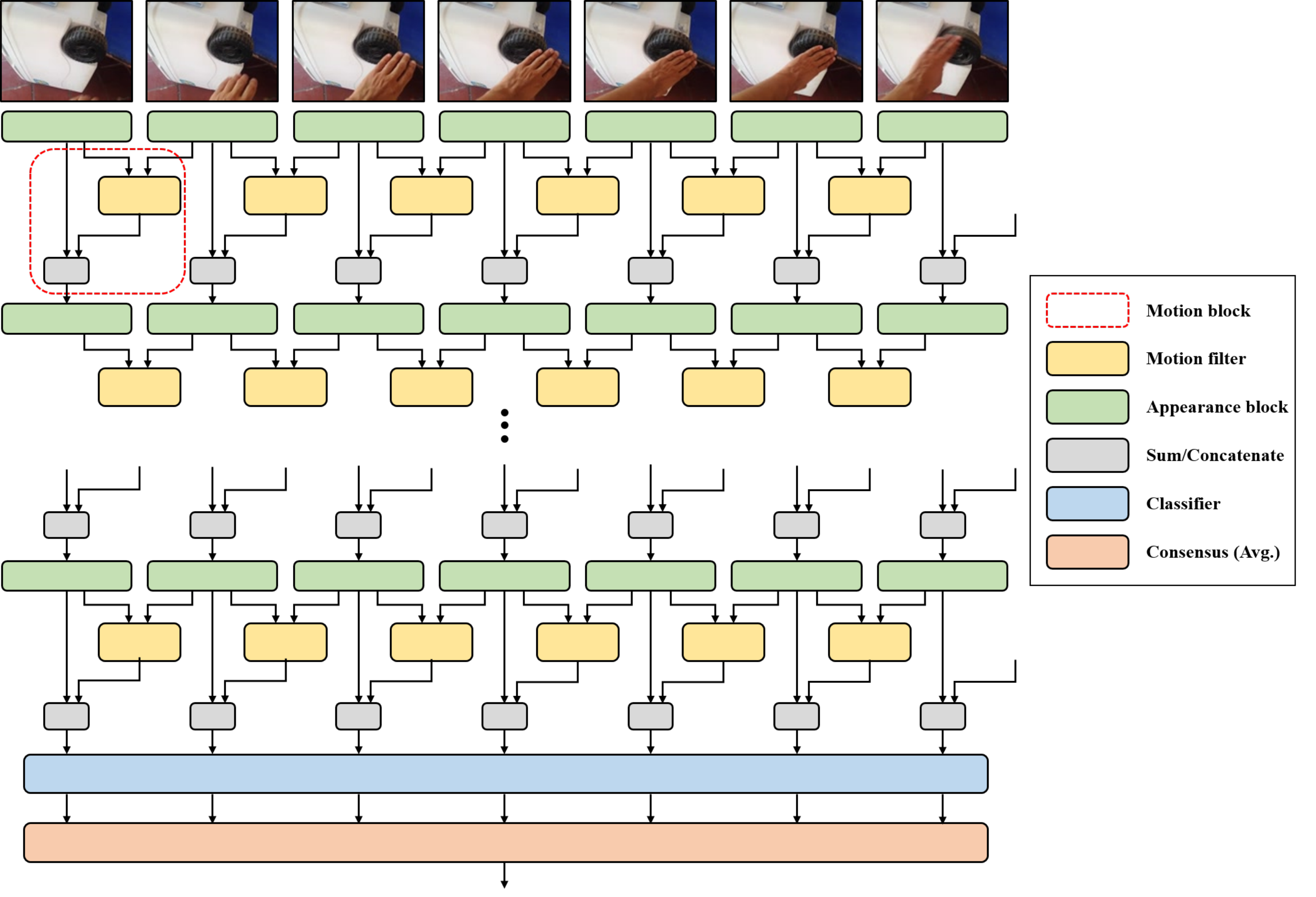}
    \caption{The overall architecture of MFNet. The proposed network is composed of appearance blocks and motion blocks which encode spatial and temporal information. A motion block takes two consecutive feature maps from respective appearance blocks and extracts spatio-temporal information with the proposed fixed motion filters. The accumulated feature maps from the appearance blocks and motion blocks are used as an input to the next layer. \mg{This figure shows the case of $K=7$.}}
    \label{fig:arch}
\end{figure}

\subsection{Motion Feature Network}

The \nj{proposed architecture of MFNet} is illustrated in Figure \ref{fig:arch}. We construct our architecture based on \textit{temporal segment network} (TSN) \cite{wang2016temporal} \mg{which works on a sequence of $K$ snippets sampled from the entire video.} Our network is composed of two major components. One is \textit{appearance block} which encodes the spatial information. \nj{This} can be \nj{any of the architectures} used in image classification tasks. In our experiments, we use ResNet~\cite{he2016deep} as our backbone \nj{network for appearance blocks}. 
Another component is \textit{motion block} which encodes temporal information. To model the motion representation, it takes two consecutive feature maps \nj{of the corresponding consecutive frames} from the same hierarchy\footnote{\nj{We use the term \textit{hierarchy} to represent the level of abstraction. A layer or a block of layers can correspond to a hierarchy.} } as inputs and then \nj{extracts} the temporal information using \nj{a set of fixed motion filters which will be} described in the next \nj{subsection.} The extracted spatial and temporal features in each hierarchy should \nj{be} properly propagated to the next hierarchy. To fully utilize two types of information, we provide several schemes to accumulate them \nj{for the next hierarchy}.

\subsection{Motion Representation}
To capture the motion representation, one \nj{of the commonly used approaches} in action recognition is using optical flow as inputs to \nj{a CNN}. Despite its important role in the action recognition tasks, optical flow is computationally expensive in practice. In order to replace the role of optical flow \nj{and to extract temporal features}, we propose motion filters which \nj{have a close relationship with the} optical flow.

\subsubsection{Approximation of Optical Flow}
To approximate the feature-level optical flow hierarchically, we propose a modular structure named motion filter. Typically, the brightness consistency constraint of optical flow is defined as follows:
\begin{equation}
 \label{eq:optical}
 {I}(x + \Delta x, y + \Delta y, t + \Delta t) = {I}(x,y,t),
\end{equation}
where ${I}(x,y,t)$ denotes the pixel \nj{value} at the location $(x,y)$ of a frame at time $t$. 
Here, $\Delta{x}$ and $\Delta{y}$ denote the spatial displacement in \nj{horizontal and vertical} axis respectively. 
\nj{The optical flow $(\Delta x, \Delta y)$ that meets (\ref{eq:optical})} 
is calculated between two consecutive image frames at time $t$ and $t+\Delta{t}$ at every location of \nj{an image}. 

Originally, solving \nj{an} optical flow problem is to find the optimal solution $(\Delta{x}^{\ast},\Delta{y}^{\ast})$ through \nj{an optimization technique}. However, it is \nj{hard} to solve (\ref{eq:optical}) directly without additional constraints \nj{such as spatial or temporal smoothness assumptions. Also, it takes much time to obtain a dense (pixelwise) optical flow.} 

In this paper, the primary goal is to find the temporal features derived from optical flow to help classifying action recognition rather than finding the optimal solution to optical flow. Thus, we \nj{extend (\ref{eq:optical})} to feature space by replacing \nj{an image} ${I}(x,y,t)$ with \nj{the corresponding feature maps} ${F}(x,y,t)$ and define a residual features ${R}$ as follows:
\begin{equation}
\label{eq:residual}
R_l(x,y,\Delta{t}) =F_l(x+\Delta{x},y+\Delta{y},t+\Delta{t}) - 
F_l(x,y,t),
\end{equation}
where $\textit{l}$ \nj{denotes the index of the layer or hierarchy}, ${F}_l$ is the \nj{$l$-th} feature maps from the basic network.
${R}$ is the residual features produced by two features from the same layer $l$. Given $\Delta{x}$ and $\Delta{y}$, the residual features $R$ can be easily calculated by subtracting two adjacent features at time $t$ and $t+\Delta{t}$. To fully utilize optical flow constraints in feature level, $R$ tends to have lower \nj{absolute} intensity. As searching for the lowest \nj{absolute} value in each location of feature map is trivial \nj{but time-consuming}, we design a set of \nj{predefined fixed} directions \nj{$\mathbb{D} = \{(\Delta{x}, \Delta{y})\}$} to restrict the search space. For convenience, \nj{in our implementation,} \mg{we \nj{restrict} $\Delta{x}, \Delta{y}\in{\{0,\pm 1\}}$ and $\left|\Delta{x}\right|+\left|\Delta{y}\right|\le 1$.} Shifting one pixel along each spatial dimension in \nj{the} image space 
\nj{is responsible for capturing a small amount of optical flow (\ie small movement), 
while one pixel in the feature space at a higher hierarchy of a CNN can capture larger optical flow (\ie large movement) as it looks at a larger receptive field.}

\begin{figure}[t]
	\centering
    \includegraphics[width=.7\linewidth]{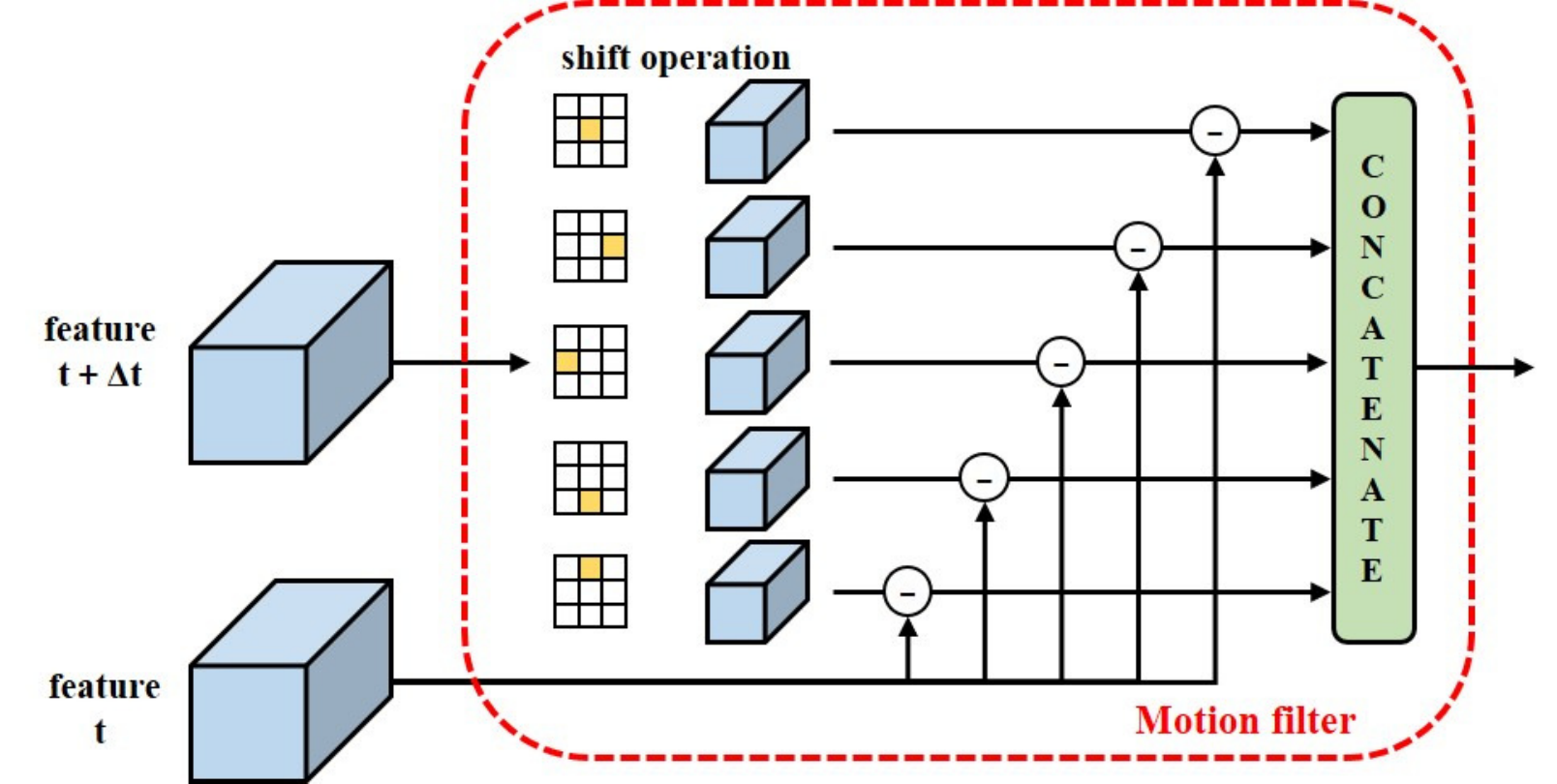}
    \caption{Motion filter.
Motion filter generates \nj{spatio-temporal} features from two consecutive feature maps. Feature \nj{map} at time $t+\Delta{t}$ is shifted by \nj{a} predefined set of fixed directions and each of them is subtracted from \nj{the} feature map at time $t$. With concatenation of features from all directions, motion filter can represent \nj{spatio-temporal} information.
    }
    \label{fig:motionfilter_shifting}
\end{figure}

\subsubsection{Motion Filter}
The motion filter is a modular structure calculated by two feature maps extracted from shared networks feed-forwarded by two consecutive frames as inputs. As shown in Figure~\ref{fig:motionfilter_shifting}, the motion filter takes features $F_l(t)$ \nj{and} $F_l(t+\Delta{t})$ at time $t$ and $t+\Delta{t}$ as inputs. \nj{The predefined} set of directions $\mathbb{D}$ is only applied to the features at time ${t+\Delta{t}}$ as illustrated Figure~\ref{fig:motionfilter_shifting}. We follow the shift operation proposed in \cite{wu2017shift}. It moves each channel of its input tensor in a different spatial direction \nj{$\delta \triangleq (\Delta{{x}}, \Delta{{y}}) \in \mathbb{D} $}. This can be alternatively done with widely used depth-wise convolution, whose kernel size is \nj{determined by} the maximum value of $\Delta{x}$ and $\Delta{y}$ in $\mathbb{D}$. \nj{For example, on our condition, $\Delta x, \Delta y \in \{0, \pm 1\}$, we can implement with $3 \times 3$ kernels as shown in Figure~\ref{fig:motionfilter_shifting}.} Formally, the shift operation can be formulated as:
\begin{align}
{G}_{k,l,m}^{\delta} & = \sum_{i,j} {K}_{i,j}^{\delta} F_{k+\hat{i},l+\hat{j},m},
\end{align}
\begin{align}
{K}_{i,j}^{\delta} & = 
  \begin{cases}
    1 & \quad \text{if } i=\Delta{x}  \text{ and }  j=\Delta{y},\\
    0 & \quad \text{otherwise.}
  \end{cases}
\end{align}

Here, \nj{the subscript indicates the index of a matrix or a tensor,}
\nj{$\delta \triangleq (\Delta x, \Delta y) \in \mathbb{D}$ is a displacement vector, $F \in{\mathbb{R}^{W\times H \times C}}$} is the input tensor and $\hat{i}=i-\lfloor W/2 \rfloor$, $\hat{j}=j-\lfloor H/2 \rfloor$ are the re-centered spatial indices ($\lfloor \cdot \rfloor$ is the floor operation). The indices $k,l$ and $i,j$ \nj{are those} along spatial dimensions and \nj{$m$ is a channel-wise index}. We get a set $\mathbb{G}$=\{$G^{\delta}_{t+\Delta t} | {\delta} \in \mathbb{D} $\}, where $G^{\delta}_{t+\Delta t}$ represents the shifted feature map \nj{by an amount of $\delta$ at time} $t+\Delta{t}$. Then, each of them is subtracted by $F_t$\footnote{For convenience, here, we use the notation $F_t$ and $G_{t+\Delta t}$ instead of $F(t)$ and $G(t+\Delta t)$. The meaning of a subscript will be obvious in the context.}. 
\nj{Because the concatenated feature map is constructed by temporal subtraction on top of the spatially shifted features, the feature map contains spatio-temporal information suitable for action recognition. As mentioned in Section \ref{sec:rel}, this is quite different from optical-flow-guided features in \cite{DBLP:journals/corr/abs-1711-11152} which use two types of feature maps obtained by  temporal subtraction and spatial Sobel filters.} \mg{Also, it is distinct from `subtractive correlation layer' in \cite{dosovitskiy2015flownet} with respect to the implementation and the goal. `Subtractive correlation layer' is utilized to find correspondences for better reconstruction, while, the proposed motion filter is aimed to encode directional information between two feature maps via learnable parameters.}

\subsection{Motion Block}

\begin{figure}[t]
	\centering
    \subfigure[Element-wise sum]{\includegraphics[width=.4\linewidth]{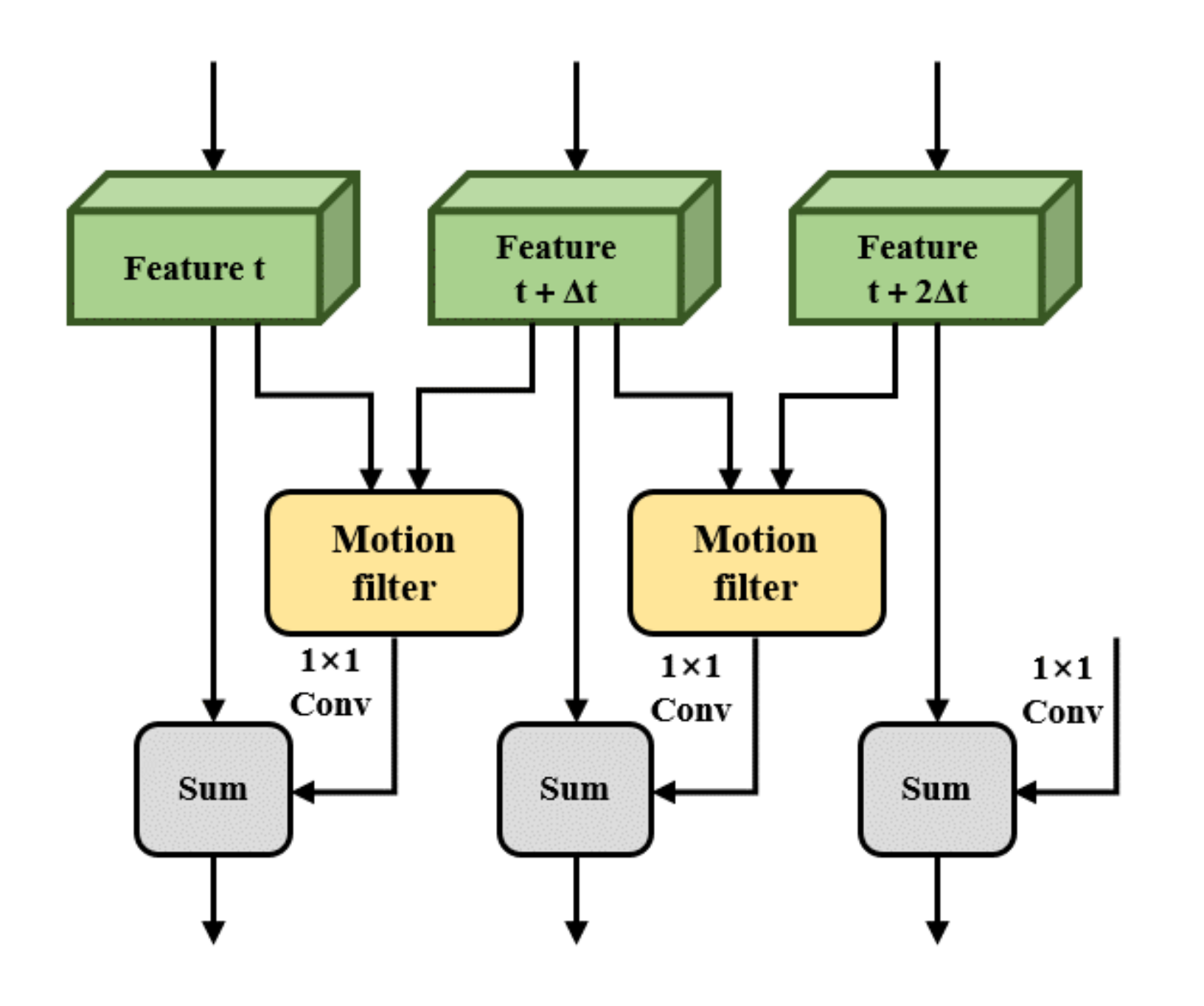}}
    \subfigure[Concatenation]{\includegraphics[width=.4\linewidth]{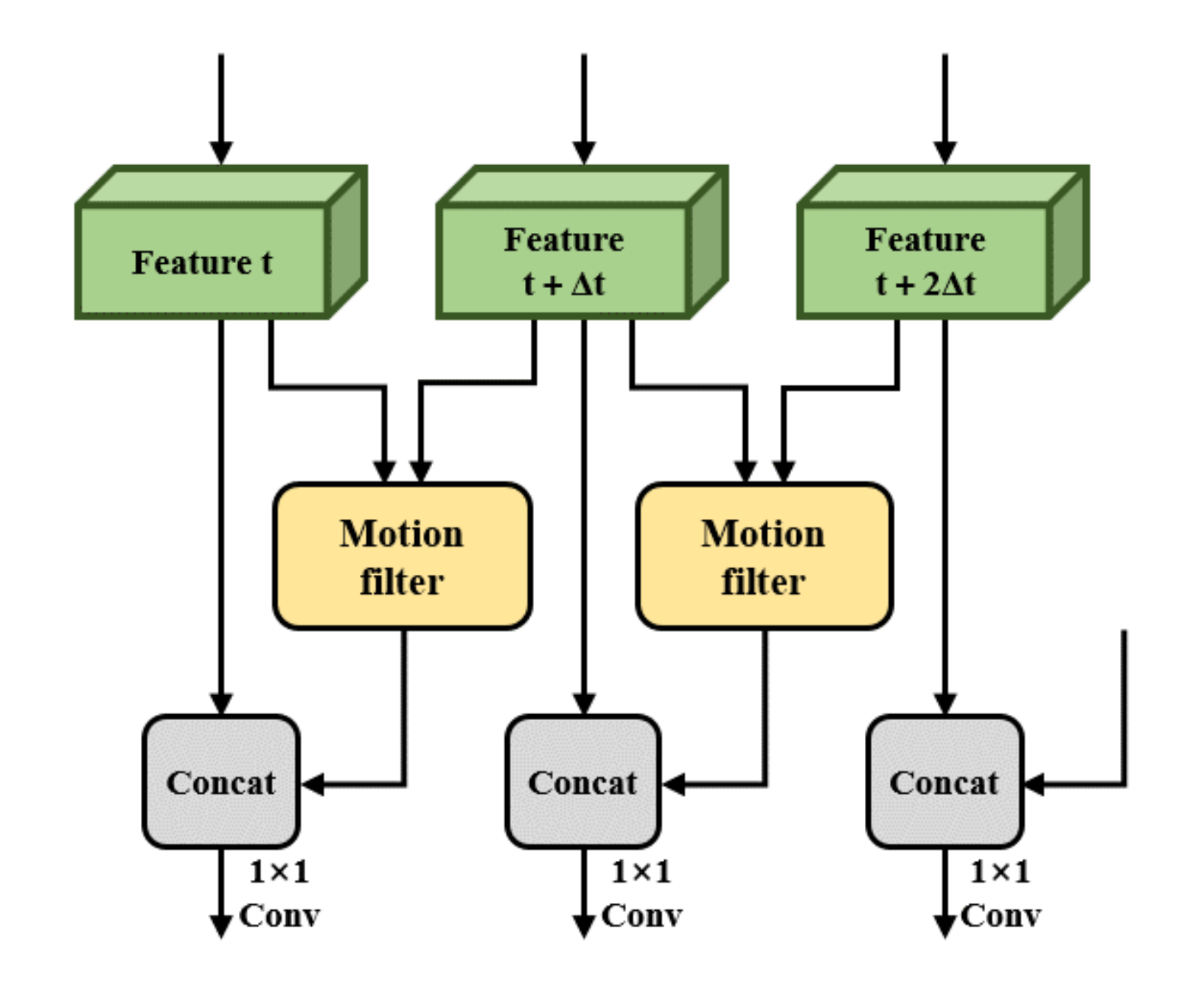}}
    \caption{Two ways to aggregate spatial and temporal information from appearance block and motion filter.}
    \label{fig:motionblock}
\end{figure}

As mentioned above, the motion filter is a modular structure \nj{which} can be adopted to any intermediate layers of two appearance blocks \nj{consecutive in time}. In order to propagate spatio-temporal information properly, we provide several building blocks. Inspired by the recent success of \mg{residual block used} in \textit{residual networks} (ResNet) in many challenging image recognition tasks, we develop a new building block named motion block to propagate spatio-temporal information between two adjacent appearance blocks \nj{into deeper layers}.

\subsubsection*{Element-wise Sum}
A simple and direct way to aggregate two different \nj{characteristics} of information is the element-wise sum operation. As illustrated in Figure ~\ref{fig:motionblock}(a), a set of \nj{motion features $R_{t}^\delta \triangleq F_t - G_{t+\Delta t}^{\delta} \in \mathbb{R}^{W\times{H}\times{C}}$, $\delta \in \mathbb{D}$}, generated by motion filter are concatenated along channel dimension \nj{to produce} a tensor \nj{${M}_{t} = [R^{\delta_1}_t | R^{\delta_2}_t | \cdots | R^{\delta_S}_t] \in{\mathbb{R}^{W\times{H}\times{N}}}$, where $[\cdot|\cdot]$ denotes a concatenation operation, $N = S \times C$ and $S$ is the number of the predefined directions in $\mathbb{D}$.} It is further compressed by $1 \times 1$ convolution filters to produce output $\hat{M}_t$ with the same dimension as $F_{t}$. Finally, the features from the appearance block $F_{t}$ and those from the motion filters $\hat{M}_t$ are summed up \nj{to produce inputs to the next hierarchy.}

\subsubsection*{Concatenation}
Another popular way to combine the \nj{appearance and the motion} features is calculated by the concatenation operation. In this paper, \nj{the motion features $M_{t}$} mentioned above are directly concatenated with each of the \nj{appearance} features $F_{t}$ as depicted in Figure~\ref{fig:motionblock}(b). \nj{A set of $1 \times 1$ convolution filters} is  also exploited to encode spatial and temporal information \nj{after the concatenation}. The $1 \times 1$ convolution reduces the channel dimension \nj{as we desire.}
It also implicitly encodes spatio-temporal features to find the relationship between two \nj{different types of features: appearance and motion features.} 

% 4. Experiments --------------------------------------------------
\section{Experiments}
\label{sec:exp}
\nj{In this section, the proposed MFNet is applied to action recognition problems and the experimental results of MFNet are compared with those of other action recognition methods. As datasets, Jester \cite{dataset_jester} and Something-Something \cite{goyal2017something} are used because these cannot be easily recognized by just seeing a frame as already mentioned in Section~\ref{sec:intro}. \nj{They also are} suitable \nj{for observing} the effectiveness of the proposed motion blocks. 
}
We \nj{also} perform comprehensive ablation studies to prove the \nj{effectiveness} of the MFNets.

\subsection{Experiment Setup}
\label{sec:expsetup}

To conduct comprehensive ablation studies on video classification tasks with motion blocks, first we describe our base network framework.

\subsubsection{Base Network Framework}
We select the TSN framework \cite{wang2016temporal} as \nj{our base} network architecture to train MFNet. TSN is an effective and efficient \nj{video processing framework for} action recognition tasks. TSN samples a sequence of frames from an entire video and \nj{aggregates} individual predictions into a video-level score. Thus, TSN framework is well suited for our motion blocks because each block directly extracts the temporal relationships between adjacent snippets \nj{in a batch manner}. 

\mg{In this paper}, we mainly choose ResNet~\cite{he2016deep} as our base network to extract spatial feature maps. For the sake of \nj{clarity}, we divide it into $6$ stages. Each stage has \nj{a} number of stacked residual blocks and each block is composed of several convolutional and batch normalization \cite{ioffe2015batch} layers with Rectified Linear Unit (ReLU) \cite{nair2010rectified} for non-linearity. \nj{The} final stage consists of a global pooling layer and a classifier. \nj{Our base network differs from} the original ResNet \nj{in that it contains the} max pooling layer in the first stage. \nj{Except this,} our base network is the same as the conventional ResNet. The backbone network can be replaced by any other network architecture and our motion blocks can be inserted into the network all in the same way \nj{regardless of the type of the network used.} 

\subsubsection{Motion Blocks}
To form MFNet, we insert our motion blocks into \nj{the} base network. In case of using ResNet, each motion block is located right after the last residual block of every stage except for the last stage (global pooling and classification layers). Then, MFNet automatically learns to represent spatio-temporal information from consecutive frames, \nj{leading} \nj{the} conventional base \nj{CNN} to extract richer information that combines both appearance and motion features. We also add \nj{an} $1 \times 1$ convolution before each motion block to reduce the number of channels. Throughout the paper, we reduce the number of input channels to motion block by a factor of $16$ with the $1\times1$ convolutional layer. We add a batch normalization layer after the $1\times1$ convolution to adjust the scale \nj{to fit to} the features in \nj{the} backbone network. 

\subsubsection{Training}
\nj{In} the datasets of Jester and Something-Something, RGB images extracted from videos at 12 frames per second with a height of $100$ pixels are provided. To augment training samples, we exploit random cropping method with scale-jittering. The width and height of \nj{a cropped image} are determined by multiplying the shorter side of the image by \nj{a scale which is} randomly selected in the set of $\{ 1.0, 0.875, 0.75, 0.625 \}$. Then \nj{the} cropped image is resized to $112\times112$, because \nj{the} width of the original images is relatively small compared to that of other datasets.
Note that we do not adopt random horizontal \nj{flipping} to \nj{the} cropped images of Jester dataset, because some classes are a symmetrical pair, such as \textit{`Swiping Left'} and \textit{`Swiping Right'}, and \textit{`Sliding Two Fingers Left'} and \textit{`Sliding Two Fingers Right'}. 

Since motion block extracts \nj{temporal motion} features from adjacent feature maps, a frame interval between frames is a very important hyper-parameter. We have trained our model with \nj{the} fixed-time sampling strategy. However, in our experiments, it leads to worse results than the random sampling strategy in \cite{wang2016temporal}. With \nj{a} random \nj{interval}, the method forces the network to learn through frames composed of various intervals. Interestingly, we get better performance on Jester and Something-Something datasets with the temporal sampling interval diversity. 

We use the stochastic gradient descent algorithm to learn network parameters. The batch size is set to $128$, the momentum is set to $0.9$ and weight decay is set to $0.0005$. All MFNets are trained from scratch \nj{and} we train our models with batch normalization layers \cite{ioffe2015batch}. \nj{The learning} rate is initialized as $0.01$ and decreases \nj{by a factor of $0.1$ for} every $50$ epochs. The training procedure \nj{stops} after $120$ epochs. To mitigate over-fitting effect, we adopt dropout~\cite{srivastava2014dropout} after the global pooling layer with \nj{a dropout} ratio of $0.5$. To speed up training, we employ a multi-GPU data-parallel strategy with 4 NVIDIA TITAN-X GPUs.

\subsubsection{Inference}
We select equi-distance $10$ frames without the random shift. We test our models on sampled frames whose image size is rescaled to $112\times112$. After that, we \nj{aggregate separate predictions from each frame and average them} before softmax normalization to get the final prediction.  

\subsection{Experimental Results}
\label{sec:experiment_results}
% TABLE ::: Number of Segments
\setlength{\tabcolsep}{4pt}
\begin{table}[t]
\begin{center}
\caption{\nj{Top-1 and Top-5 classification accuracies for different networks with different numbers of training segments ($3,5, 7$).} The compared networks are TSN baseline, MFNet concatenation version (MFNet-C), and MFNet element-wise sum version (MFNet-S) on Jester and Something-Something validation sets. All models use ResNet-50 as a backbone network and are trained from scratch.}
\begin{tabular}{c|c|c|c|c|c}
\hline
\multicolumn{2}{c|}{Dataset} & \multicolumn{2}{c|}{Jester} & \multicolumn{2}{c}{Something-Something} \\
\hline
\multicolumn{1}{c|}{model} & \multicolumn{1}{c|}{\textit{K}} & top-1 acc. & top-5 acc. & top-1 acc. & top-5 acc.\\
\hline
\multirow{3}*{Baseline}
& 3  & 82.4\% & 98.9\% &  6.6\% & 21.5\% \\
& 5  & 82.8\% & 98.9\% &  9.8\% & 28.6\% \\
& 7  & 81.0\% & 98.5\% &  8.1\% & 24.7\% \\
\hline
\multirow{3}*{MFNet-C50}
& 3  & 90.4\% & 99.5\% & 17.4\% & 42.6\% \\
& 5  & 95.1\% & 99.7\% & 31.5\% & 61.9\% \\
& 7  & 96.1\% & 99.7\% & 37.3\% & 67.2\% \\
\hline
\multirow{3}*{MFNet-S50}
& 3  & 91.0\% & 99.6\% & 15.4\% & 39.2\% \\
& 5  & 95.6\% & 99.8\% & 28.7\% & 59.1\% \\
& 7  & 96.3\% & 99.8\% & 37.1\% & 67.8\% \\
\hline
\end{tabular}
\label{table:valid_number_of_segments}
\end{center}
\end{table}
\setlength{\tabcolsep}{1.4pt}
% TABLE ::: Number of Segments
The Jester\cite{dataset_jester} is a crowd-acted video dataset for generic human hand gestures recognition. It consists $118,562$ videos for training, $14,787$ videos for validation, and $14,743$ videos for testing. 
The Something-Something\cite{goyal2017something} is also a crowd-acted densely labeled video dataset of basic human interactions with daily objects. It contains $86,017$ videos for training, $11,522$ videos for validation, and $10,960$ videos for testing. Each of both datasets is for the action classification task involving $27$ and $174$ human action categories respectively. 
We report validation results of our models on the validation sets, and test results from the official leaderboards\footnote{https://www.twentybn.com/datasets/jester}$^{, }$\footnote{https://www.twentybn.com/datasets/something-something}. 

\subsubsection{Evaluation on The Number of Segments}
Due to the nature of our MFNet, the number of segments, $K$, \nj{in the training} is one of the important parameters. Table~\ref{table:valid_number_of_segments} shows the comparison results of different models while changing the number of segments from $3$ to $7$ with the same evaluation strategies. 
We observe that as the number of segments increases, the performance of overall models increases. \nj{The} performance of the MFNet-C50 (which means that MFNet concatenate version with ResNet-50 as a backbone network) with $7$ segments is by far the better than \nj{the same network} with $3$ segments: $96.1\%$ vs. $90.4\%$ and $37.3\%$ vs. $17.4\%$ on Jester and Something-Something datasets respectively. \nj{The trend is the same for MFNet-S50, the network with element-wise sum.} Also, unlike baseline \nj{TSN}, MFNets show significant performance improvement as the number of segments increases from $3$ to $5$. 

These improvements imply that increasing $K$ reduces the interval between sampled frames which allows our model to extract richer information. Interestingly, MFNet-S achieves slightly higher top-1 accuracy($0.2\%$ to $0.6\%$) than MFNet-C on Jester dataset, and MFNet-C shows better performance($0.2\%$ to $2.8\%$) than MFNet-S on Something-Something dataset. On the other hand, because the TSN baseline is learned from scratch, performance was worse than expected. It can be seen \nj{that} TSN spatial model without pre-training barely generates any \nj{action-related visual features in Something-Something dataset.
}

\subsubsection{Comparisons of Network Depths}
% TABLE ::: Network Depths
\setlength{\tabcolsep}{4pt}
\begin{table}[t]
\begin{center}
\caption{
\nj{Top-1 and Top-5 classification accuracies for different depths of MFNet's base network. ResNet\cite{he2016deep} is used as the base network.} 
The values are on JESTER and Something-Something validation sets. All models are trained from scratch, with $10$ segments.}
\begin{tabular}{c|l|c|c|c|c}
\hline
\multicolumn{2}{c|}{Dataset} & \multicolumn{2}{c|}{Jester} & \multicolumn{2}{c}{Something-Something} \\
\hline
\multicolumn{1}{c|}{model} & \multicolumn{1}{c|}{backbone} & top-1 acc. & top-5 acc. & top-1 acc. & top-5 acc. \\
\hline
\multirow{4}*{MFNet-C}
& ResNet-18  	& 96.3\% & 99.8\% & 39.4\% & 69.1\% \\
& ResNet-50  	& 96.6\% & 99.8\% & 40.3\% & 70.9\% \\
& ResNet-101 	& 96.7\% & 99.8\% & 43.9\% & 73.1\% \\
& ResNet-152 	& 96.5\% & 99.8\% & 43.0\% & 73.2\% \\
\hline
\end{tabular}
\label{table:valid_network_depth}
\end{center}
\end{table}
\setlength{\tabcolsep}{1.4pt}
% TABLE ::: Network Depths
Table~\ref{table:valid_network_depth} compares \nj{the performances as the depths of MFNet's backbone network changes}. \nj{In the table, we can see}  that MFNet-C with ResNet-18 achieves comparable performance as the 101-layered ResNet using almost $76\%$ fewer parameters ($11.68$M vs. $50.23$M). It is generally known that as \nj{CNNs} become deeper, more features can be expressed~\cite{he2016deep,simonyan2014very,szegedy2015going}. However, one can see that because most of the videos in Jester dataset are composed of almost similar kinds of human appearances,  \nj{the static visual} entities are very little related to action classes. Therefore, the network depth does not appear to have a significant effect on performance. In Something-Something case, accuracy gets also saturated. It could be explained that \nj{generalization of a model} seems to be difficult without pre-trained weights on other large-scale datasets, such as \nj{Imagenet \cite{deng2009imagenet} and Kinetics \cite{kay2017kinetics}}.

\subsubsection{Comparisons with I3D}
% TABLE ::: Comparisons with I3D ------------------------------------------------------------------------------------------------------------------
\setlength{\tabcolsep}{4pt}
\begin{table}[t]
\begin{center}
\caption{Top-1 and Top-5 accuracies for MFNet-C and I3D with different backbone networks (ResNet-18 and ResNet-101) on Jester and Something-Something datasets. We re-implemented the I3D model \cite{carreira2017quo}. LR means initial learning rate.}
\begin{tabular}{c|c|c|c|c|c|c}
\hline
\multicolumn{3}{c|}{Dataset} & \multicolumn{2}{c|}{Jester} & \multicolumn{2}{c}{Something-Something} \\
\hline
\multicolumn{1}{c|}{model} & \multicolumn{1}{c|}{$K$} & \multicolumn{1}{c|}{LR} & top-1 acc. & top-5 acc. & top-1 acc. & top-5 acc. \\
\hline
\multirow{1}*{MFNet-C18}
& 10 & 0.01 & 96.3\% & 99.8\% & 39.4\% & 69.1\% \\  % 11.68M
\multirow{1}*{MFNet-C101}
& 10 & 0.01 & 96.7\% & 99.8\% & 43.9\% & 73.1\% \\  % 50.23M
% & ResNet-50  	& 10 & 0.01 & 40.3\% & 70.9\% \\
% & ResNet-101 	& 10 & 0.01 & 43.9\% & 73.1\% \\  % 50.23M
% & ResNet-152 	& 10 & 0.01 & 43.0\% & 73.2\% \\
\hline
\multirow{4}*{I3D \cite{carreira2017quo}}
& 16 & 0.1  & 0\% & 0\% & 23.7\% & 51.1\% \\  % 12.47M  
& 32 & 0.1  & 0\% & 0\% & 38.2\% & 69.9\% \\  % 12.47M
& 16 & 0.01 & 0\% & 0\% & 34.6\% & 64.7\% \\  % 12.47M
& 32 & 0.01 & 0\% & 0\% & 43.6\% & 73.8\% \\  % 12.47M 
\hline
\end{tabular}
\label{table:I3D}
\end{center}
\end{table}
\setlength{\tabcolsep}{1.4pt}
% TABLE ::: Comparisons with I3D ------------------------------------------------------------------------------------------------------------------
We trained the I3D model \cite{carreira2017quo} from scratch for fair comparison on Something-Something dataset. We followed data augmentation methods in \cite{carreira2017quo} except for the input number of frames. Due to the lack and difference of the number of frames in Jester and Something-Something datasets, we trained I3D with 16 and 32-frame video snippets. However, at test time, we process all video frames as the same as \cite{carreira2017quo}. Also we initialize learning rates of $0.1$ and $0.01$ for a hyper-parameter search on two datasets. Other experimental setup adopts the default settings in our paper. Table~\ref{table:I3D} shows that our MFNet-C achieves competitive performance to the I3D. It is worth noting that for the validation set, we evaluate MFNet-C with sparsely sampled $10$ snippets while I3D densely predicts action classes using all the frames. 

\subsubsection{Comparisons with The State-of-the-art}
% TABLE ::: Jester & Something Validation Accuracy -----------------------------------------------------------------------------------------------
\setlength{\tabcolsep}{4pt}
\begin{table}[t]
\begin{center}
\caption{\nj{Comparison of the top-1 and top-5 validation results of various methods} on Jester and Something-something datasets. $K$ denotes the number of training segments. \nj{The results of other models are from their respective papers.}}
\begin{tabular}{l|c|c|c|c}
\hline
\multicolumn{1}{c|}{Dataset} & \multicolumn{2}{c|}{Jester} & \multicolumn{2}{c}{Something-Something} \\
\hline
\multicolumn{1}{c|}{model} & top-1 acc.& top-5 acc.& top-1 acc.& top-5 acc.\\
\hline
Pre-3D CNN + Avg\cite{goyal2017something}			& -		  & -       & 11.5\%  & 30.0\%  \\
MultiScale TRN\cite{zhou2017temporal}  				& 93.70\% & 99.59\% & 33.01\% & 61.27\% \\
MultiScale TRN (10-crop)\cite{zhou2017temporal}  	& 95.31\% & 99.86\% & 34.44\% & 63.20\% \\
\hline
MFNet-C50, $K=7$									& 96.13\% & 99.65\% & 37.31\% & 67.23\% \\
MFNet-S50, $K=7$									& 96.31\% & 99.80\% & 37.09\% & 67.78\% \\
MFNet-C50, $K=10$									& 96.56\% & 99.82\% & 40.30\% & 70.93\% \\
MFNet-S50, $K=10$									& 96.50\% & 99.86\% & 39.83\% & 70.19\% \\
MFNet-C101, $K=10$ 									& 96.68\% & 99.84\% & 43.92\% & 73.12\% \\
\hline
\end{tabular}
\label{table:valid_performance}
\end{center}
\end{table}
\setlength{\tabcolsep}{1.4pt}
% TABLE ::: Jester & Something Validation Accuracy -----------------------------------------------------------------------------------------------

% TABLE ::: Jester & Something test Accuracy -----------------------------------------------------------------------------------------------
\setlength{\tabcolsep}{4pt}
\begin{table}[t]
\begin{center}
\caption{Selected test results on the Jester and Something-Something datasets from the official leaderboards. Since the test results are continuously updated, some results that \nj{are} not reported or whose description \nj{is missing} are excluded. The complete list of test results is available on official public leaderboards. Our results are based on ResNet-101 with $K=10$ and trained from scratch. For submissions, we use the same evaluation strategies \nj{as} the validation mode.}
\begin{tabular}{l|c|l|c}
\hline
\multicolumn{2}{c|}{Jester} & \multicolumn{2}{c}{Something-Something} \\
\hline
\multicolumn{1}{c|}{model} & \multicolumn{1}{c|}{top-1 acc.} & \multicolumn{1}{c|}{model} & \multicolumn{1}{c}{top-1 acc.} \\
\hline
BesNet (from \cite{zhou2017temporal}) 	&94.23\%&BesNet (from \cite{zhou2017temporal}) 		&31.66\%\\
MultiScale TRN \cite{zhou2017temporal}	&94.78\%&MultiScale TRN \cite{zhou2017temporal} 	  &33.60\%\\
\hline
MFNet-C101 (ours)	 &96.22\%&MFNet-C101 (ours)			&37.48\%\\
\hline
\end{tabular}
\label{table:test_performance}
\end{center}
\end{table}
\setlength{\tabcolsep}{1.4pt}
% TABLE ::: Jester & Something test Accuracy -----------------------------------------------------------------------------------------------

Table~\ref{table:valid_performance} shows the top-1 and top-5 results on the validation set. Our models outperform Pre-3D CNN + Avg \cite{goyal2017something} and the MultiScale TRN \cite{zhou2017temporal}. Because Jester and Something-Something are recently released datasets in the action recognition research field, we also report the test results on the official leaderboards for each dataset for comparison with previous studies. Table~\ref{table:test_performance} shows that MFNet achieves comparable performance \nj{to the state-of-the-art methods} with $96.22\%$ and $37.48\%$ top-1 accuracies on Jester and Something-Something test datasets respectively on official leaderboards. Note that we do not introduce any other modalities, ensemble methods \nj{or} pre-trained initialization weights on large-scale datasets such as ImageNet \cite{deng2009imagenet} and Kinetics \cite{kay2017kinetics}. We only utilize officially provided RGB images as the input of our final results. Also, without 3D ConvNets and additional complex testing strategies, our method provides competitive performances on the Jester and Something-Something datasets.   

% FIGURE ::: Confusion Matrix Analysis
\begin{figure}[t]
\centering
\includegraphics[width=\linewidth]{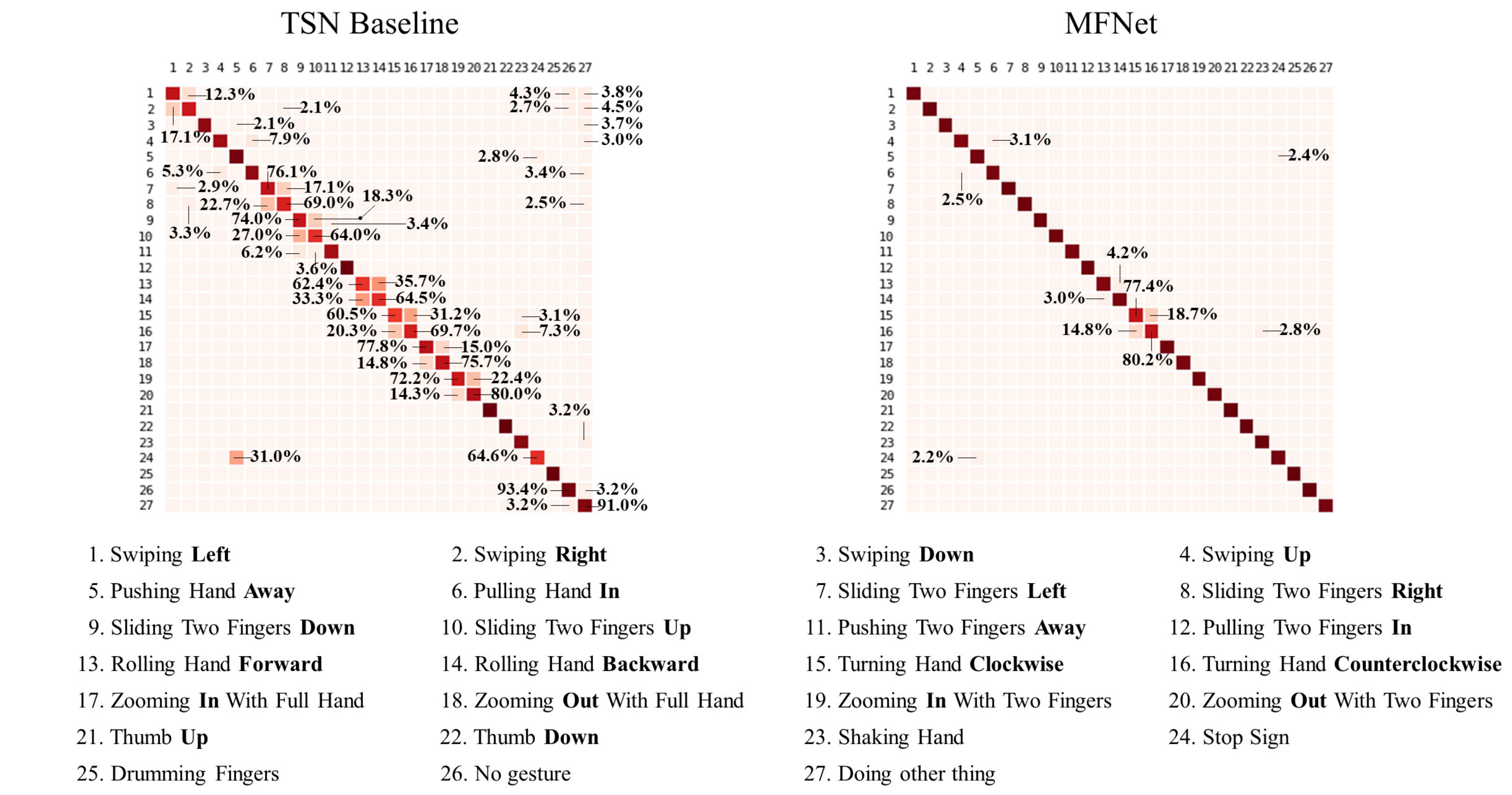}
\caption{Confusion matrices of TSN baseline and our proposed MFNet on Jester dataset. The figure is best viewed in an electronic form.} 
\label{confusion}
\end{figure}
% FIGURE ::: Confusion Matrix Analysis

% FIGURE ::: Validation Accuracy vs. Inference samples 
\begin{figure}[tb]
\centering
\begin{minipage}[t]{0.45\linewidth}
	\subfigure[Jester]{\includegraphics[width=\linewidth]{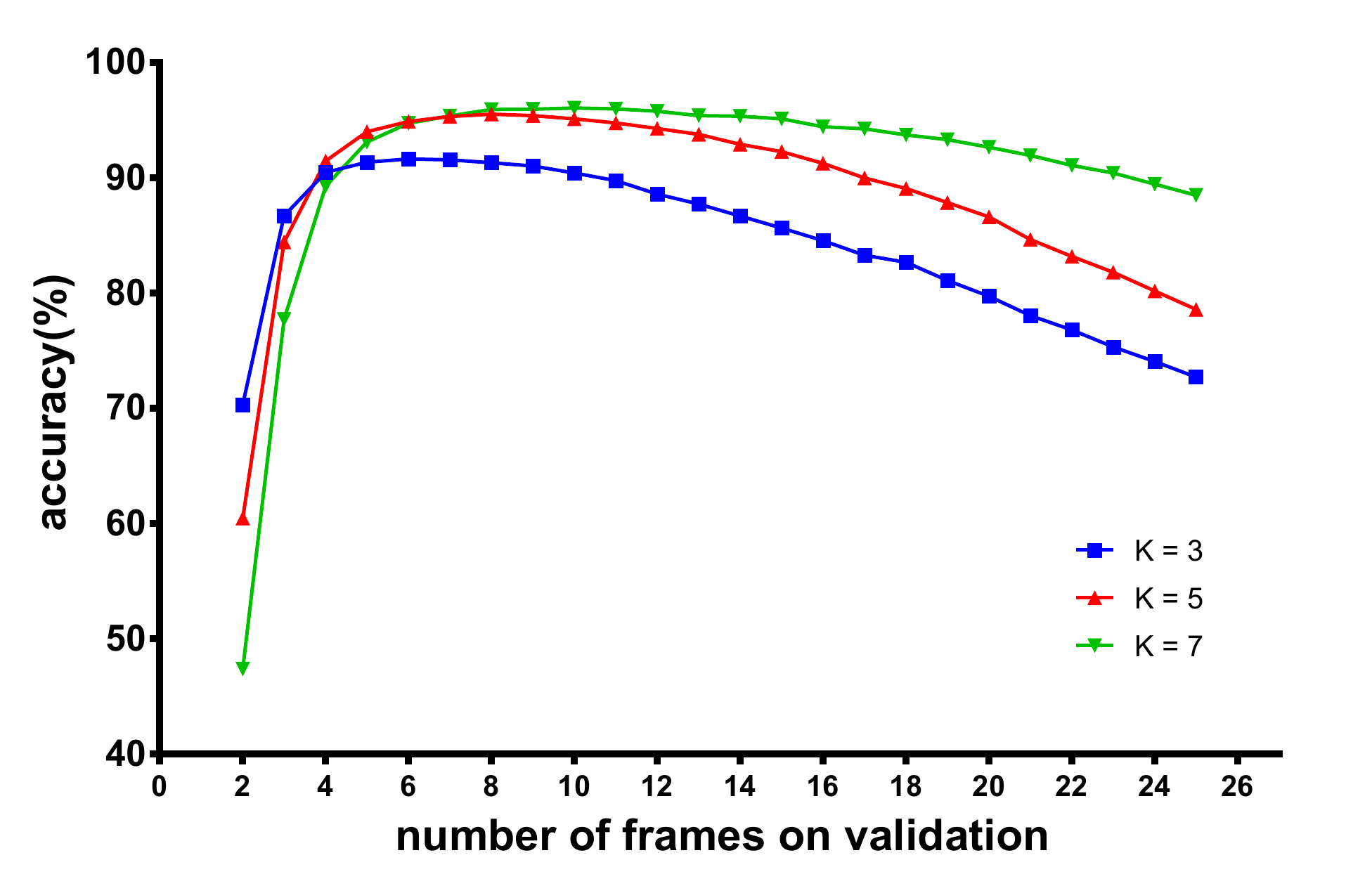}}
    \label{f3}
\end{minipage}
\begin{minipage}[t]{0.45\linewidth}
	\subfigure[Something-Something]{\includegraphics[width=\linewidth]{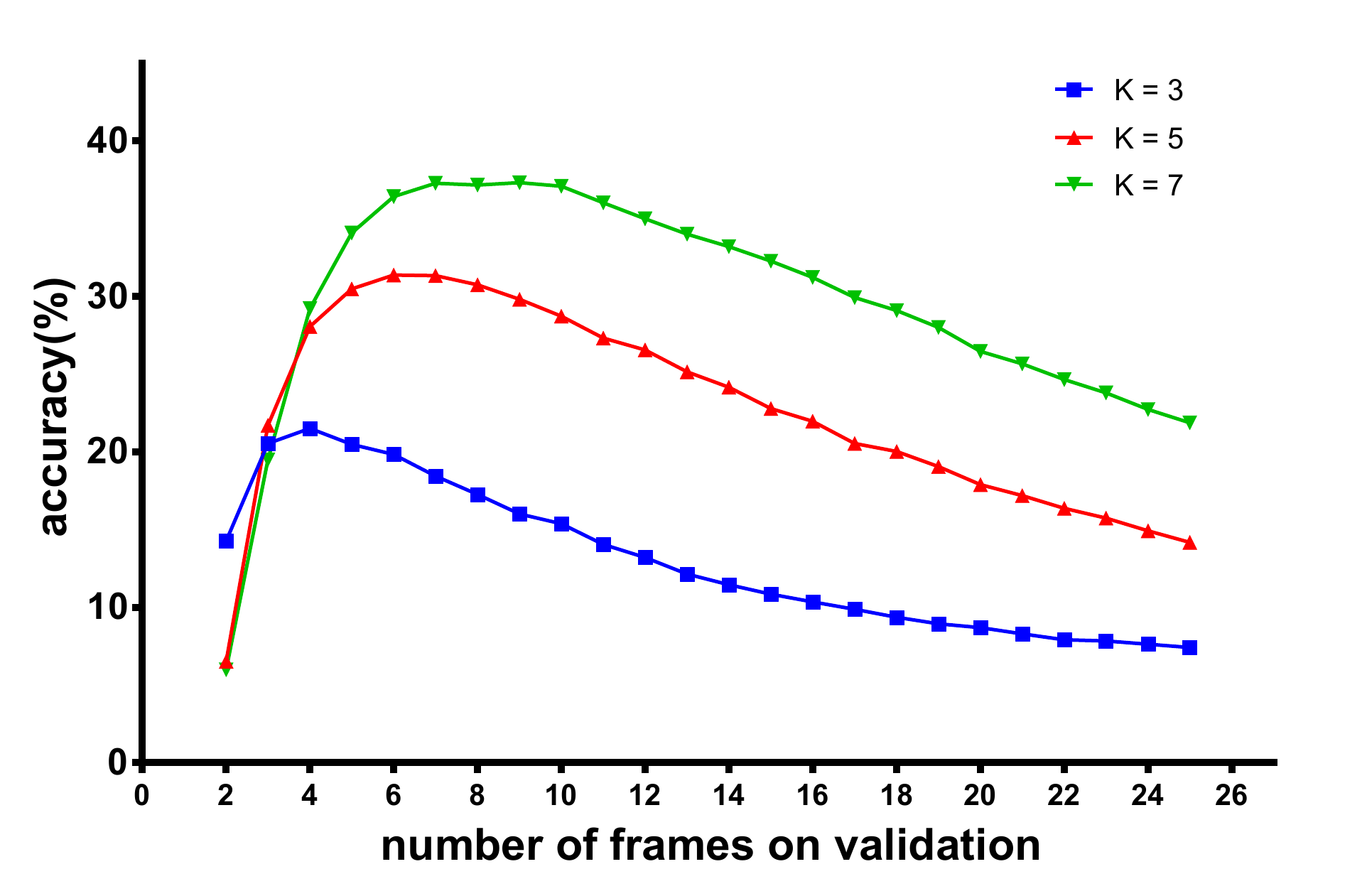}}
    \label{f4}
\end{minipage}
\caption{Validation accuracies trained with the different number of segments $K$, while varying the number of validation segments from $2$ to $25$. The x-axis \nj{represents} the number of segments at inference time and the y-axis \nj{is} the validation accuracy of the MFNet-C50 trained with different $K$.}
\label{figure:varying_inference}
\end{figure}

% Analysis on behavior of MFNet
\subsection{Analysis on The Behavior of MFNet}
\subsubsection{Confusion Matrix}
We analyze the effectiveness of MFNet comparing with the baseline. Figure~\ref{confusion} shows the confusion matrices of TSN baseline (left) and MFNet (right) on Jester dataset. Class numbers and the corresponding class names are listed below. Figure~\ref{confusion} suggests that the baseline model confuses one action class with its counterpart class. That is, it has trouble classifying temporally symmetric action pairs. For example, (\textit{`Swiping Left', `Swiping Right'}) and (\textit{`Two Finger Down', `Two Finger Up'}) are temporally symmetric pairs. 

In case of baseline, it predicts an action class by simply averaging the results of sampled frames. Consequently, if there is no optical flow information, it might fail to distinguish some temporal symmetric action pairs. Specifically, we get 62.38\% accuracy on \textit{`Rolling Hand Forward'} class among 35.7\% of which is misclassified as \textit{`Rolling Hand Backward'}. 
In contrast, our MFNet showed significant improvement over baseline model as shown in Figure~\ref{confusion} (right). In our experiments, we get the accuracy of 94.62\% on \textit{`Rolling Hand Forward'} class among 4.2\% of which is identified as \textit{`Rolling Hand Backward'}. It proves the ability of MFNet \nj{in capturing} the motion representation.

\subsubsection{Varying Number of Segments in The Validation Phase}
We evaluated the models which have different numbers of frames in \nj{the} inference phase. Figure~\ref{figure:varying_inference} shows the \nj{experimental} results of MFNet-C50 \nj{on} Jester (left) and Something-Something (right) datasets. As discussed in Section~\ref{sec:experiment_results}, \nj{$K$, the number of segments in the training phase is a} crucial parameter on performance. As we can see, overall performance for all \nj{the} number of validation segments is superior on large $K$ (7). Meanwhile, the optimal number of validation segments for each $K$ is different. Interestingly, it does not coincide with $K$ but is slightly larger than $K$.
Using more segments reduces the frame interval which allows extracting more precise spatio-temporal features. It brings the effect of improving performance. However, it does not last \nj{if the numbers in the training and the validation phases differ much}.

% 5. Conclusions --------------------------------------------------
\section{Conclusions}
\label{sec:con}
In this paper, we present MFNet, a unified network containing appearance blocks and motion blocks which can represent both spatial and temporal information for action recognition \nj{problems}. Especially, we propose the motion filter that outputs the \nj{motion} features by performing the shift operation with the fixed set of predefined \nj{directional filters and subtracting the resultant feature maps from the feature maps} of \nj{the preceding} frame. This module can be attached to any existing CNN-based network with \nj{a} small additional cost. We evaluate our model on two datasets, Jester and Something-Something, \nj{and obtain outperforming results compared to the existing results by training the network from scratch in an end-to-end manner}. Also, we perform comprehensive ablation studies and analysis on the behavior of MFNet to show the effectiveness of our method. In the future, we will validate our network on large-scale action recognition dataset and additionally investigate the usefulness of the \nj{proposed} motion block. 

\bibliographystyle{splncs04}
\bibliography{egbib}

\begin{thebibliography}{10}
\providecommand{\url}[1]{\texttt{#1}}
\providecommand{\urlprefix}{URL }
\providecommand{\doi}[1]{https://doi.org/#1}

\bibitem{dataset_jester}
The 20bn-jester dataset. \url{https://www.twentybn.com/datasets/jester}

\bibitem{carreira2017quo}
Carreira, J., Zisserman, A.: Quo vadis, action recognition? a new model and the
  kinetics dataset. In: 2017 IEEE Conference on Computer Vision and Pattern
  Recognition (CVPR). pp. 4724--4733. IEEE (2017)

\bibitem{dalal2006human}
Dalal, N., Triggs, B., Schmid, C.: Human detection using oriented histograms of
  flow and appearance. In: European conference on computer vision. pp.
  428--441. Springer (2006)

\bibitem{deng2009imagenet}
Deng, J., Dong, W., Socher, R., Li, L.J., Li, K., Fei-Fei, L.: Imagenet: A
  large-scale hierarchical image database. In: Computer Vision and Pattern
  Recognition, 2009. CVPR 2009. IEEE Conference on. pp. 248--255. IEEE (2009)

\bibitem{dosovitskiy2015flownet}
Dosovitskiy, A., Fischer, P., Ilg, E., Hausser, P., Hazirbas, C., Golkov, V.,
  Van Der~Smagt, P., Cremers, D., Brox, T.: Flownet: Learning optical flow with
  convolutional networks. In: Proceedings of the IEEE International Conference
  on Computer Vision. pp. 2758--2766 (2015)

\bibitem{feichtenhofer2016spatiotemporal}
Feichtenhofer, C., Pinz, A., Wildes, R.: Spatiotemporal residual networks for
  video action recognition. In: Advances in neural information processing
  systems. pp. 3468--3476 (2016)

\bibitem{feichtenhofer2016convolutional}
Feichtenhofer, C., Pinz, A., Zisserman, A.: Convolutional two-stream network
  fusion for video action recognition  (2016)

\bibitem{goyal2017something}
Goyal, R., Kahou, S.E., Michalski, V., Materzynska, J., Westphal, S., Kim, H.,
  Haenel, V., Fruend, I., Yianilos, P., Mueller-Freitag, M., et~al.: The”
  something something” video database for learning and evaluating visual
  common sense. In: Proc. ICCV (2017)

\bibitem{hara2017learning}
Hara, K., Kataoka, H., Satoh, Y.: Learning spatio-temporal features with 3d
  residual networks for action recognition. In: Proceedings of the ICCV
  Workshop on Action, Gesture, and Emotion Recognition. vol.~2, p.~4 (2017)

\bibitem{he2016deep}
He, K., Zhang, X., Ren, S., Sun, J.: Deep residual learning for image
  recognition. In: Proceedings of the IEEE conference on computer vision and
  pattern recognition. pp. 770--778 (2016)

\bibitem{ioffe2015batch}
Ioffe, S., Szegedy, C.: Batch normalization: Accelerating deep network training
  by reducing internal covariate shift. In: International conference on machine
  learning. pp. 448--456 (2015)

\bibitem{jiang2014thumos}
Jiang, Y., Liu, J., Zamir, A.R., Toderici, G., Laptev, I., Shah, M.,
  Sukthankar, R.: Thumos challenge: Action recognition with a large number of
  classes (2014)

\bibitem{karpathy2014large}
Karpathy, A., Toderici, G., Shetty, S., Leung, T., Sukthankar, R., Fei-Fei, L.:
  Large-scale video classification with convolutional neural networks. In:
  Proceedings of the IEEE conference on Computer Vision and Pattern
  Recognition. pp. 1725--1732 (2014)

\bibitem{kay2017kinetics}
Kay, W., Carreira, J., Simonyan, K., Zhang, B., Hillier, C., Vijayanarasimhan,
  S., Viola, F., Green, T., Back, T., Natsev, P., et~al.: The kinetics human
  action video dataset. arXiv preprint arXiv:1705.06950  (2017)

\bibitem{Kuehne11}
Kuehne, H., Jhuang, H., Garrote, E., Poggio, T., Serre, T.: {HMDB}: a large
  video database for human motion recognition. In: Proceedings of the
  International Conference on Computer Vision (ICCV) (2011)

\bibitem{kuehne2013hmdb51}
Kuehne, H., Jhuang, H., Stiefelhagen, R., Serre, T.: Hmdb51: A large video
  database for human motion recognition. In: High Performance Computing in
  Science and Engineering ‘12, pp. 571--582. Springer (2013)

\bibitem{lecun1995convolutional}
LeCun, Y., Bengio, Y., et~al.: Convolutional networks for images, speech, and
  time series. The handbook of brain theory and neural networks
  \textbf{3361}(10), ~1995 (1995)

\bibitem{miech2017learnable}
Miech, A., Laptev, I., Sivic, J.: Learnable pooling with context gating for
  video classification. arXiv preprint arXiv:1706.06905  (2017)

\bibitem{nair2010rectified}
Nair, V., Hinton, G.E.: Rectified linear units improve restricted boltzmann
  machines. In: Proceedings of the 27th international conference on machine
  learning (ICML-10). pp. 807--814 (2010)

\bibitem{ng2016actionflownet}
Ng, J.Y.H., Choi, J., Neumann, J., Davis, L.S.: Actionflownet: Learning motion
  representation for action recognition. arXiv preprint arXiv:1612.03052
  (2016)

\bibitem{ng2015beyond}
Ng, J.Y.H., Hausknecht, M., Vijayanarasimhan, S., Vinyals, O., Monga, R.,
  Toderici, G.: Beyond short snippets: Deep networks for video classification.
  In: Computer Vision and Pattern Recognition (CVPR), 2015 IEEE Conference on.
  pp. 4694--4702. IEEE (2015)

\bibitem{simonyan2014two}
Simonyan, K., Zisserman, A.: Two-stream convolutional networks for action
  recognition in videos. In: Advances in neural information processing systems.
  pp. 568--576 (2014)

\bibitem{simonyan2014very}
Simonyan, K., Zisserman, A.: Very deep convolutional networks for large-scale
  image recognition. arXiv preprint arXiv:1409.1556  (2014)

\bibitem{soomro2012ucf101}
Soomro, K., Zamir, A.R., Shah, M.: Ucf101: A dataset of 101 human actions
  classes from videos in the wild. arXiv preprint arXiv:1212.0402  (2012)

\bibitem{srivastava2014dropout}
Srivastava, N., Hinton, G., Krizhevsky, A., Sutskever, I., Salakhutdinov, R.:
  Dropout: A simple way to prevent neural networks from overfitting. The
  Journal of Machine Learning Research  \textbf{15}(1),  1929--1958 (2014)

\bibitem{DBLP:journals/corr/abs-1711-11152}
Sun, S., Kuang, Z., Ouyang, W., Sheng, L., Zhang, W.: Optical flow guided
  feature: {A} fast and robust motion representation for video action
  recognition. CoRR  \textbf{abs/1711.11152} (2017),
  \url{http://arxiv.org/abs/1711.11152}

\bibitem{szegedy2015going}
Szegedy, C., Liu, W., Jia, Y., Sermanet, P., Reed, S., Anguelov, D., Erhan, D.,
  Vanhoucke, V., Rabinovich, A., et~al.: Going deeper with convolutions. Cvpr
  (2015)

\bibitem{tran2015learning}
Tran, D., Bourdev, L., Fergus, R., Torresani, L., Paluri, M.: Learning
  spatiotemporal features with 3d convolutional networks. In: Computer Vision
  (ICCV), 2015 IEEE International Conference on. pp. 4489--4497. IEEE (2015)

\bibitem{tran2017convnet}
Tran, D., Ray, J., Shou, Z., Chang, S.F., Paluri, M.: Convnet architecture
  search for spatiotemporal feature learning. arXiv preprint arXiv:1708.05038
  (2017)

\bibitem{wang2011action}
Wang, H., Kl{\"a}ser, A., Schmid, C., Liu, C.L.: Action recognition by dense
  trajectories. In: Computer Vision and Pattern Recognition (CVPR), 2011 IEEE
  Conference on. pp. 3169--3176. IEEE (2011)

\bibitem{wang2013action}
Wang, H., Schmid, C.: Action recognition with improved trajectories. In:
  Computer Vision (ICCV), 2013 IEEE International Conference on. pp.
  3551--3558. IEEE (2013)

\bibitem{wang2017appearance}
Wang, L., Li, W., Li, W., Van~Gool, L.: Appearance-and-relation networks for
  video classification. arXiv preprint arXiv:1711.09125  (2017)

\bibitem{wang2016temporal}
Wang, L., Xiong, Y., Wang, Z., Qiao, Y., Lin, D., Tang, X., Van~Gool, L.:
  Temporal segment networks: Towards good practices for deep action
  recognition. In: European Conference on Computer Vision. pp. 20--36. Springer
  (2016)

\bibitem{wu2017shift}
Wu, B., Wan, A., Yue, X., Jin, P., Zhao, S., Golmant, N., Gholaminejad, A.,
  Gonzalez, J., Keutzer, K.: Shift: A zero flop, zero parameter alternative to
  spatial convolutions. arXiv preprint arXiv:1711.08141  (2017)

\bibitem{wu2015fusing}
Wu, Z., Jiang, Y.G., Wang, X., Ye, H., Xue, X., Wang, J.: Fusing multi-stream
  deep networks for video classification. arXiv preprint arXiv:1509.06086
  (2015)

\bibitem{zhang2016real}
Zhang, B., Wang, L., Wang, Z., Qiao, Y., Wang, H.: Real-time action recognition
  with enhanced motion vector cnns. In: Computer Vision and Pattern Recognition
  (CVPR), 2016 IEEE Conference on. pp. 2718--2726. IEEE (2016)

\bibitem{zhou2017temporal}
Zhou, B., Andonian, A., Torralba, A.: Temporal relational reasoning in videos.
  arXiv preprint arXiv:1711.08496  (2017)

\bibitem{zhu2017hidden}
Zhu, Y., Lan, Z., Newsam, S., Hauptmann, A.G.: Hidden two-stream convolutional
  networks for action recognition. arXiv preprint arXiv:1704.00389  (2017)

\end{thebibliography}
\end{document}